\documentclass[10pt,twocolumn,letterpaper]{article}

\usepackage{cvpr}              % To produce the CAMERA-READY version
\usepackage{algorithm}
\usepackage{algorithmic}
\usepackage{graphicx}
\usepackage{amsmath}
\usepackage{amssymb}
\usepackage{booktabs}
\usepackage{amsfonts}

\usepackage{multirow}
\usepackage{array}
\usepackage{color}
\usepackage{subcaption}
\usepackage{colortbl}
\usepackage{arydshln}
\graphicspath{{images/}}
\usepackage{bbding}
\usepackage{makecell}
\usepackage{hhline}
\usepackage{bm}

\usepackage[pagebackref,breaklinks,colorlinks]{hyperref}

\usepackage[capitalize]{cleveref}
\crefname{section}{Sec.}{Secs.}
\Crefname{section}{Section}{Sections}
\Crefname{table}{Table}{Tables}
\crefname{table}{Tab.}{Tabs.}

\def\cvprPaperID{*****} % *** Enter the CVPR Paper ID here
\def\confName{CVPR}
\def\confYear{2022}

\begin{document}

%%%%%%%%% TITLE - PLEASE UPDATE
% \title{1st Place Solutions to Out of Vocabulary Scene Text Understanding Challenge}
\title{Runner-Up Solution to ECCV 2022 Challenge on Out of Vocabulary Scene Text Understanding: Cropped Word Recognition}

\author{Zhangzi Zhu, Yu Hao, Wenqing Zhang, Chuhui Xue, Song Bai\\
ByteDance Inc.
}
\maketitle

%%%%%%%%% ABSTRACT
\begin{abstract}
%Scene text recognition has attracted increasing interests in recent years due to its wide range of applications in multilingual translation, autonomous driving, \emph{etc}. In this report, we describe our solution to the Out of Vocabulary Scene Text Understanding (OOV-ST) Challenge, which aims to extract out-of-vocabulary (OOV) words from natural scene images. For the Cropped Word Text Recognition Track, we use SCATTER and provide effective solutions. Finally, we achieve 69.73\% word accuracy in IV+OOV words, ranking 1st in the Cropped Word Recognition challenge of ECCV2022 TiE Workshop.

This report presents our 2nd place solution to ECCV 2022 challenge on Out-of-Vocabulary Scene Text Understanding (OOV-ST) : Cropped Word Recognition. This challenge is held in the context of ECCV 2022 workshop on Text in Everything (TiE), which aims to extract out-of-vocabulary words from natural scene images. In the competition, we first pre-train SCATTER on the synthetic datasets, then fine-tune the model on the training set with data augmentations. Meanwhile, two additional models are trained specifically for long and vertical texts. Finally, we combine the output from different models with different layers, different backbones, and different seeds as the final results. Our solution achieves a word accuracy of 59.45\% when considering out-of-vocabulary words only.
\end{abstract}

\section{Introduction}
% introduction to scene text recognition
% why need to study on oov text recognition
%Texts in scenes contain rich semantic information that is very important and valuable in many practical applications such as multilingual translation, autonomous driving, \emph{etc}. With the development of deep learning, many researchers explore the use of language information to rectify the text recognition results on in-vocabulary (IV) words if they have been appeared in the training set. However, the language information of out-of-vocabulary (OOV) words are usually difficult to learn if they have never been seen during training. 

This report focuses on scene text recognition, which aims to recognize a sequence of characters from a cropped image. The task is evaluated in both in-vocabulary (IV) and out-of-vocabulary (OOV) words, where ``IV" refers to the text instances has been seen in the training set and ``OOV" means the text instances are unseen accordingly.

%ECCV 2022 Challenge on Out of Vocabulary Scene Text Understanding aims to evaluate the model performances on recognizing OOV words. In this competition, the training, validation and test sets are composed of several commonly used datasets, including ICDAR13~\cite{karatzas2013icdar}, ICDAR15 \cite{karatzas2015icdar}, MLT19 \cite{nayef2019icdar2019}, COCO-Text~\cite{veit2016coco}, TextOCR~\cite{singh2021textocr}, HierText~\cite{long2022towards}, and OpenImagesText~\cite{krylov2021open}. Additionally, participants are allowed to generate synthetic data with a provided dictionary of the 90k most frequent English words. Since the competition puts more emphasis on OOV instances but do not want to drop IV words completely, two leaderboards are recognized: (1) on OOV words only; (2) on both IV and OOV words by averaging the IV and OOV scores.

ECCV 2022 challenge on Out-of-Vocabulary Scene Text Understanding (OOV-ST) \footnote{\scriptsize\url{https://rrc.cvc.uab.es/?ch=19}}, held together with ECCV 2022 workshop on Text in Everything (TiE) \footnote{\scriptsize\url{https://sites.google.com/view/tie-eccv2022/challenge}}, favors the recognition of OOV words. In this competition, the training, validation and test sets are composed of several commonly used datasets, including ICDAR13~\cite{karatzas2013icdar}, ICDAR15~\cite{karatzas2015icdar}, MLT19~\cite{nayef2019icdar2019}, COCO-Text~\cite{veit2016coco}, TextOCR~\cite{singh2021textocr}, HierText~\cite{long2022towards}, and OpenImagesText~\cite{krylov2021open}. Additionally, participants are allowed to generate synthetic data with a provided dictionary of the 90k most frequent English words. Since the competition emphasizes on OOV instances but does not wish to drop IV words completely, two leaderboards are recognized, including 1) on OOV words only and 2) on both IV and OOV words by averaging the IV and OOV scores.

% our method
%In this report, we describe our solution to the Cropped Word Recognition track. We first pre-train SCATTER on the synthetic datasets which are generated from a provided dictionary of the 90k most frequent English words, and then fine-tune the model on the training set with data augmentation. In order to recognize long texts and vertical texts more accurately, extra models are trained especially for these two kinds of texts. Finally, we combine results from different layers of SCATTER, different backbones, and different seeds together. Our method achieves 69.73\% in word accuracy and ranks 1st on Cropped Word Recognition of IV+OOV words.

In this report, we describe our solution to the Cropped Word Recognition track in this challenge. Our solution achieves 59.45\% in word accuracy when considering out-of-vocabulary words only, which ranked 2nd in the competition. Details are given below.

\section{Methods}
% In this track, participants are expected to localize and recognize all the text in the image. We first pre-train our Deformable ResNet-101 and VAN-Large by using oCLIP on the training set. We then train TESTR, PAN and Mask TextSpotter with different backbones by using the pre-trained model. Finally, we combine results from different methods, different backbones, and different scales together.

%In this track, participants need to recognize OOV words from the cropped images. We try different recognition models including ASTER~\cite{shi2018aster}, ABINet~\cite{fang2021read} and SCATTER~\cite{litman2020scatter}, and finally choose SCATTER as our baseline model for its superior performance in this task. The following improvements are made:

Our solution is based on SCATTER~\cite{litman2020scatter}, with the following improvements:

\subsection{Synthetic Data}
To improve the generalization of the model, we generate 60M synthetic data with the provided dictionary of the 90k most frequent English words. We pre-train SCATTER on the synthetic data to accelerate the model convergence.

\subsection{Data Augmentation}
In both pre-training stage and fine-tuning stage, multiple data augmentations are employed. Specifically, we follow ABINet~\cite{fang2021read} and use geometry transformation (\emph{i.e.}, rotation, affine and perspective transformations), image quality deterioration (\emph{i.e.}, gaussian noise, motion blur and JPEG compression), color jitter, \emph{etc}.

\subsection{Long Texts and Vertical Texts}
It is observed that the baseline model does not perform well over long texts and vertical texts, which is caused by sample imbalance in the training set. Since the proportion of long texts and vertical texts is too small, models tend to focus more on regular texts. Therefore, we train two more SCATTER-based models especially for better recognition of long texts and vertical texts.

\vspace{1ex}\noindent\textbf{Long Texts.} We consider images with large aspect ratios (\emph{i.e.} larger than 9:1) as long images and texts in such images are denoted as long texts. We set the maximum length of texts in long-text model to 50 and train on long images only.

\vspace{1ex}\noindent\textbf{Vertical Texts.} Images with small aspect ratios (\emph{i.e.} less than 1:3) are considered as vertical images and texts in these images are denoted as vertical texts. We additionally train a SCATTER-based model on these vertical images only. Since the directions of texts in these images are undetermined, we randomly rotate them by 90 degrees clockwise (or counterclockwise) during the training phase. In the testing stage, they are uniformly rotated 90 degrees clockwise and passed to the vertical-text model.

\vspace{1ex}\noindent\textbf{Inference.} In the inference stage, images with different aspect ratios are passed into different models. If the aspect ratio of input image is less than 1:3 (or larger than 9:1), it is sent into the vertical-text (or long-text) model. The rest images are passed into the baseline models for inference.
%During the inference stage, images with aspect ratio smaller than 1:3, between 1:3 and 9:1, and greater than 9:1 are passed to vertical-text models, baseline models, and long-text models, respectively.

% 

\subsection{Ensemble}
\noindent\textbf{Internal Ensemble.}
According to~\cite{litman2020scatter}, the output sequence of characters is from the final selective decoder during inference, but the training loss consists of all five selective decoder blocks with the same weight. Therefore, we combine internal results by averaging the output probabilities of all five blocks at each time step.

\vspace{1ex}\noindent\textbf{External Ensemble.}
We train SCATTER models with different backbones and different seeds, and finally combine their outputs together as the way in internal ensemble.

\begin{table}[tb]
\centering
\resizebox{0.49\textwidth}{!}{ 
\begin{tabular}{ccccccc}
\toprule
\textbf{SCATTER} & \textbf{DA} & \textbf{Synthetic} & \textbf{Long\&Vertical} & \textbf{CRW} & \textbf{mED} \\
\midrule
{$ \checkmark $} & & & & 54.7 & 1.48 \\
{$ \checkmark $} & {$ \checkmark $} & & & 57.0 & 1.38 \\
{$ \checkmark $} & {$ \checkmark $} & {$ \checkmark $} & & 59.3 & 1.36 \\
{$ \checkmark $} & {$ \checkmark $} & {$ \checkmark $} & {$ \checkmark $} & \textbf{60.0} &  \textbf{1.31} \\
\bottomrule
\end{tabular}
}
\caption{Ablation results of Data Augmentation (DA), Synthetic Data (Synthetic), Long texts and Vertical texts (Long\&Vertical) in the OOV validation set.}
\label{Table1}
\end{table}

\begin{table}[tb]
\centering
\small
\begin{tabular}{llcc}
\toprule
\textbf{Image Type} & \textbf{Backbone} & \textbf{CRW} & \textbf{mED} \\
\midrule
Grayscale images &  ResNet18  & 60.0 & 1.31 \\
RGB images & ResNet50 & \textbf{61.8} & \textbf{1.26} \\
\bottomrule
\end{tabular}
\caption{Evaluation of image types and backbones in the OOV validation set.}
\label{Table5}
\end{table}

\begin{table}[tb]
\centering
\small
\begin{tabular}{cccccc}
\toprule
\textbf{SCATTER} & \textbf{Internal} & \textbf{External} & \textbf{CRW} & \textbf{mED} \\
\midrule
{$ \checkmark $} & & & 61.8 & 1.26 \\
{$ \checkmark $} & {$ \checkmark $} & & 62.0 & 1.21 \\
{$ \checkmark $} & {$ \checkmark $} & {$ \checkmark $} & \textbf{63.9} & \textbf{1.17} \\
\bottomrule
\end{tabular}
\caption{Evaluation of internal and external ensemble in the OOV validation set.}
\label{Table2}
\end{table}

% \begin{table}[tb]
% 	\begin{center}
% 	\setlength{\tabcolsep 5.4pt}{
% 		\begin{tabular}{cccc}
% 		    \hline
% 			\multicolumn{2}{c}{\textbf{Ensemble}} & \multicolumn{2}{c}{\textbf{Metric}} \\
%     		\cmidrule(r){1-2}\cmidrule(l){3-4}
%     		\textbf{Internal} & \textbf{External} & \textbf{CRW} & \textbf{mED} \\ \hline
%     		 &  & 61.8 & 1.26 \\ 
% 			{$ \checkmark $} &  & 62.0 & 1.21 \\
% 			{$ \checkmark $} & {$ \checkmark $} & 63.9 & 1.17 \\ \hline
% 		\end{tabular}}
% 	\end{center}
% 	\caption{Evaluation of internal and external ensemble in the OOV validation set.}
% \label{Table2}\end{table}

\begin{table}[tb]
	\begin{center}
\resizebox{0.49\textwidth}{!}{ 
		\begin{tabular}{lccccc}
		    \toprule
			 \multirow{2}{*}{\textbf{Method}} & \multicolumn{2}{c}{\textbf{IV}} & \multicolumn{2}{c}{\textbf{OOV}} & \textbf{IV\&OOV} \\
			     \cmidrule(r){2-3}\cmidrule(lr){4-5} \cmidrule(l){6-6}
    		& \textbf{CRW} & \textbf{ED} & \textbf{CRW} & \textbf{ED} & \textbf{CRW} \\ 
    					\midrule
    	Our solution & 79.7 & 113482 & 59.5 & 43890 & 69.6 \\  
    		\bottomrule
		\end{tabular}}
	\end{center}
	\caption{Final results in the test set over IV, OOV, and IV\&OOV words .}
\label{Table3}
\end{table}

\section{Experiments}
\subsection{Evaluation Metrics}
The results are evaluated in terms of two metrics, ~\emph{i.e.,} Correcly Recognized Words (CRW) and Edit Distance (ED)\footnote{\url{https://rrc.cvc.uab.es/?ch=19&com=tasks}}. If not specified otherwise, all the ablation studies are done in the out-of-vocabulary (OOV) validation set.

\subsection{Experiment Results}
Table \ref{Table1} shows the ablation results of using data augmentation, synthetic data, and long\&vertical texts. In this study, we take grayscale images as input and choose ResNet18 as backbone. As Table \ref{Table1} shows, both data augmentation and synthetic data bring an improvement by 2.3\% in Correcly Recognized Words (CRW), validating that they can enhance the generalization of the model. By assembling two extra models specifically for long and vertical texts, CRW is improved by 0.7\% overall in the validation set, even though long and vertical texts only account for only 5\% in the validation set.

%The improvement is still encouraging especially considering long and vertical texts account for nearly 5\% in the validation dataset.

We also tried different image types and backbones. As shown in Table~\ref{Table5}, it leads to an improvement of 1.8\% in terms of CRW by using RGB images and ResNet-50 as the backbone. 

Furthermore, Table~\ref{Table2} demonstrates the effectiveness of two ensemble strategies. As Table~\ref{Table2} shows, we obtain an improvement of 0.2\% in terms of CRW by internally combining all five blocks together. Additionally, external ensemble improves the CRW score from 62.0\% to 63.9\%.

The final results of IV, OOV, IV+OOV words in the test set are shown in Table \ref{Table3}. Moreover, we combine our recognizer with our oCLIP-based~\cite{xue2022language} text detector and obtain the best performance in the end-to-end recognition track for OOV words.

% Since stronger backbone (\emph{i.e. ResNet50}) and taking RGB images as input can achieve higher word accuracy \ref{Table5}, we conduct ensemble experiments conditioned on the above new settings.
\section{Conclusion}
This report summarizes the details of our runner-up solution to ECCV 2022 Challenge on Out of Vocabulary Scene Text Understanding in the track of Cropped Word Recognition.

%%%%%%%%% REFERENCES
{\small
\bibliographystyle{ieee_fullname}
\bibliography{egbib}
}

\end{document}